\documentclass[runningheads]{llncs}

\usepackage{eccv}
\usepackage{multirow}

\usepackage{eccvabbrv}
\usepackage{tcolorbox}
\usepackage{graphicx}
\usepackage{booktabs}
\usepackage[accsupp]{axessibility}  

\usepackage{hyperref}

\usepackage{orcidlink}

\begin{document}

\title{Visual Text Generation in the Wild} 





\author{
Yuanzhi Zhu\inst{1}\thanks{Equal contribution. $\dagger$ Corresponding author.}\orcidlink{0000-0003-0900-1379} \and
Jiawei Liu\inst{2}$^{\star}$\orcidlink{0000-0002-6100-4000} \and
Feiyu Gao\inst{1}$^{\star}$\orcidlink{0009-0009-3206-5347} \and
Wenyu Liu\inst{2}$^{\dagger}$\orcidlink{0000-0002-4582-7488} \and
Xinggang Wang\inst{2}\orcidlink{0000-0001-6732-7823} \and
Peng Wang\inst{1}\orcidlink{0009-0001-8617-1550} \and
Fei Huang\inst{1}\orcidlink{0000-0002-3709-5053} \and
Cong Yao\inst{1}\orcidlink{0000-0001-6564-4796} \and
Zhibo Yang\inst{1}$^{\dagger}$\orcidlink{0000-0003-2343-7750}}

\institute{Alibaba Group \and Huazhong University of Science and Technology \\
\email{\{zyz.kpillow,wdp0072012,yangzhibo450,yaocong2010\}@gmail.com, \\
\{feiyu.gfy,f.huang\}@alibaba-inc.com, \{jiaweiliu,xgwang,liuwy\}@hust.edu.cn}}

\authorrunning{Y. Zhu et al.}

\maketitle

\begin{abstract}
Recently, with the rapid advancements of generative models, the field of visual text generation has witnessed significant progress. However, it is still challenging to render high-quality text images in real-world scenarios, as three critical criteria should be satisfied: (1) Fidelity:  the generated text images should be photo-realistic and the contents are expected to be the same as specified in the given conditions; (2) Reasonability: the regions and contents of the generated text should cohere with the scene; (3) Utility: the generated text images can facilitate related tasks (\eg, text detection and recognition). Upon investigation, we find that existing methods, either rendering-based or diffusion-based, can hardly meet all these aspects simultaneously, limiting their application range. Therefore, we propose in this paper a visual text generator (termed SceneVTG), which can produce high-quality text images in the wild. Following a two-stage paradigm, SceneVTG leverages a Multimodal Large Language Model to recommend reasonable text regions and contents across multiple scales and levels, which are used by a conditional diffusion model as conditions to generate text images. Extensive experiments demonstrate that the proposed SceneVTG significantly outperforms traditional rendering-based methods and recent diffusion-based methods in terms of fidelity and reasonability. Besides, the generated images provide superior utility for tasks involving text detection and text recognition. Code and datasets are available at \href{https://github.com/AlibabaResearch/AdvancedLiterateMachinery/tree/main/OCR/SceneVTG}{\textcolor{magenta}{AdvancedLiterateMachinery}}. 

\keywords{Visual text generation \and Real-world scenarios \and Conditional diffusion models }

\end{abstract}

\section{Introduction}
\label{sec:intro}
Scene text is one of the most challenging research objects within the visual-text-related research domain due to its complexity and diversity. 
However, since the data-thirsty and domain-specific characteristics of models in the deep learning era, it is extremely expensive and hard, if not impossible, to obtain sufficient real-world data. 
In various scene text tasks, numerous studies have demonstrated the importance of synthetic data~\cite{zhou2017east,shi2016crnn,fang2021abinet,liu2020abcnet}, making visual text generation~\cite{gupta2016synthetic,zhan2018verisimilar,long2020unrealtext,tang2023scene,chen2023textdiffuser,chen2023textdiffuser2,yang2024glyphcontrol,tuo2023anytext} a popular research topic. However, generating high-quality scene text images is challenging as it requires simultaneously considering three critical criteria as shown in~\cref{fig:fig1}:

\begin{figure}[t]
\centering
\includegraphics[width=0.97\textwidth]{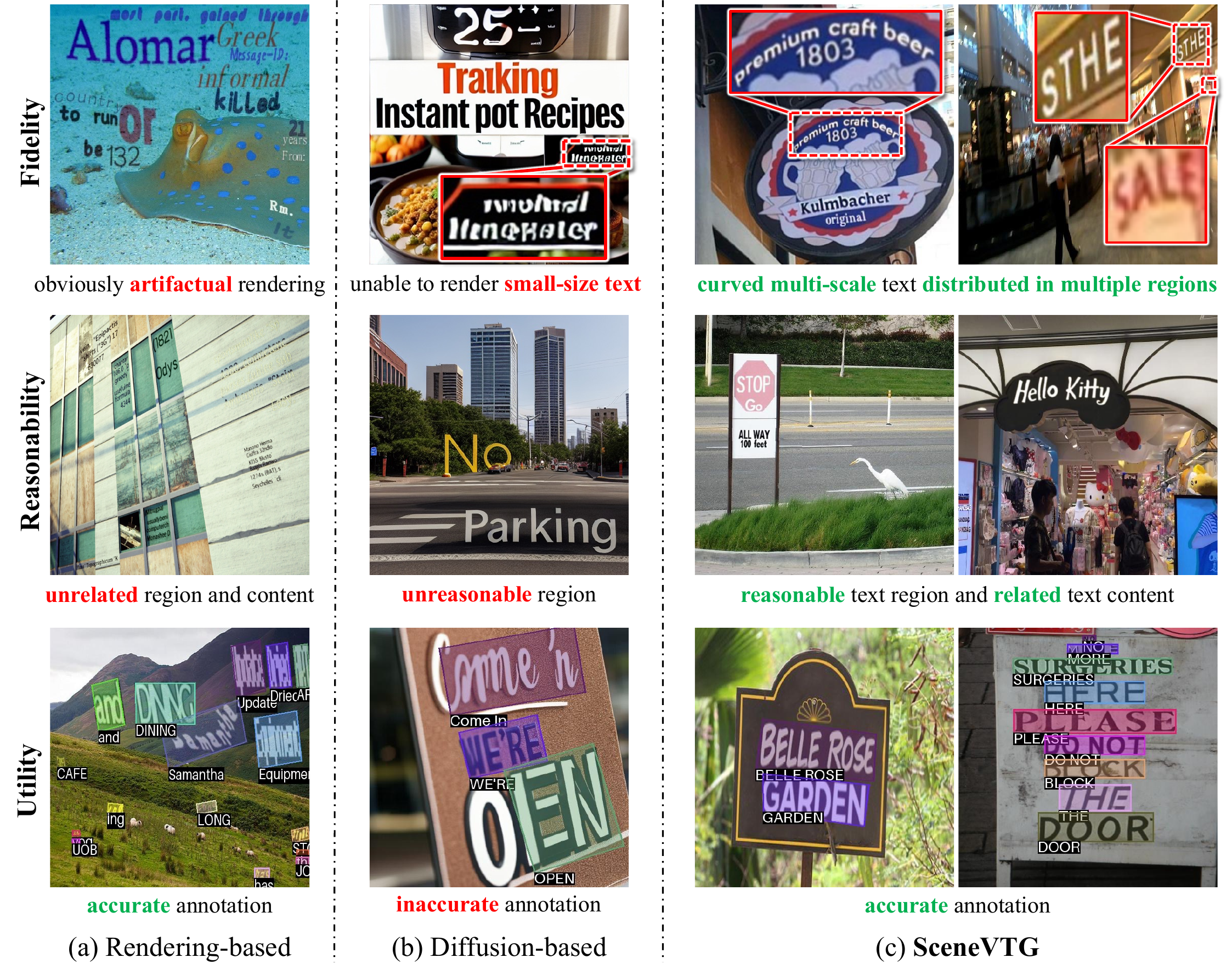}
\caption{Comparison with rendering-based methods, diffusion-based methods, and our method SceneVTG regarding fidelity, reasonability, and utility. Their \textcolor{Green}{\textbf{advances}} and \textcolor{Red}{\textbf{limitations}} are indicated in \textcolor{Green}{\textbf{green}} and \textcolor{Red}{\textbf{red}}. Zoom in for better views. }
\label{fig:fig1}
\end{figure}

\begin{itemize}
\item \textbf{Fidelity}: The generated text pixels should be integrated with the image background, with no discernible artifacts. And the text content should match the specified conditions, with no erroneous characters or superfluous text. 

\item \textbf{Reasonability}: The regions and contents of the generated text should be consistent with the image context and should not be nonsensical. 

\item \textbf{Utility}: The generated images should better enhance the performance of related tasks, such as text detection and recognition.
\end{itemize}

\begin{figure}[t]
\centering
\includegraphics[width=0.92\textwidth]{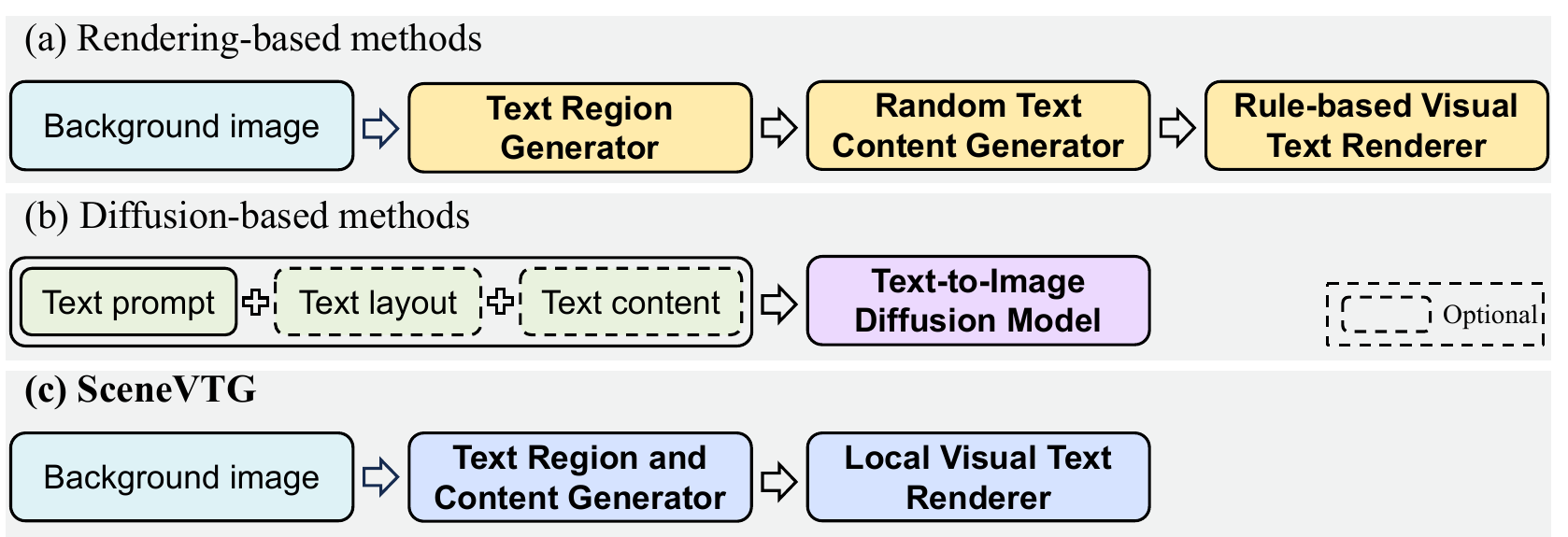}
\caption{Pipelines of rendering-based methods, diffusion-based methods, and SceneVTG.} 
\label{fig:fig2}
\end{figure}

In the early, visual text generation was primarily based on rendering and followed three steps, \ie, identifying suitable text regions via various feature maps, randomly selecting text contents from the corpus, and applying manually defined rules across various materials (\eg, fonts and colors) to render text onto the background images, as presented in~\cref{fig:fig2}~(a). 
These methods are effective for OCR model training, but they lack fidelity and reasonability due to the following problems: (1) The identified text regions might be rough and imprecise. (2) The randomly selected text contents lack correlation with each other or with the image context, causing the visual text to be nonsensical. (3) Random selection of text attributes results in text pixels that exhibit obvious artifacts.

Recently, diffusion models have demonstrated state-of-the-art fidelity and diversity in image generation, and there have been some cutting-edge works~\cite{chen2023textdiffuser,yang2024glyphcontrol,chen2023textdiffuser2,tuo2023anytext} exploring their capabilities in visual text generation. 
As shown in~\cref{fig:fig2}~(b), diffusion-based visual text generation generally generates images directly through prompt text or additionally combines text-related priors, such as text layout, content, \etc. 
Owing to the semantic alignment ability of prompt text, the foreground texts and background images are more integrated than rendering-based methods. However, existing methods demonstrate limited diversity in terms of text region and text scale, as they exhibit a strong propensity to generate text in constrained regions~\cite{chen2023textdiffuser2} and encounter difficulties in generating smaller-scale characters~\cite{tuo2023anytext} following the VAE encoding and latent diffusion paradigm. Furthermore, their utility is compromised by inaccurate text annotations, as the generated texts often fail to align with the regions indicated in the conditions and superfluous texts may also be generated outside of the specified regions.

In this paper, we explore visual text generation in real-world scenes and propose a new method called \textbf{SceneVTG}. 
As shown in~\cref{fig:fig2}~(c), SceneVTG follows a two-stage paradigm and consists of a Text Region and Content Generator~(TRCG) and a Local Visual Text Renderer~(LVTR). 
In TRCG, we leverage the visual reasoning ability of Multimodal Large Language Models (MLLMs) to identify suitable text regions on the background image across multiple scales and recommend contents that are contextually coherent and visually fitting.
Additionally, TRCG can generate curved text regions and provide accurate text annotations.
In LVTR, we utilize a local conditional diffusion model rather than relying on latent features of the whole image, enabling text generation at arbitrary scales. 
Specifically, based on the output of TRCG, LVTR constructs image-level and embedding-level conditions to generate local text regions with consistent backgrounds and then embed them into the entire image to finally achieve realistic scene text image generation. 
To train our models, we contribute a new dataset \textbf{SceneVTG-Erase}, which contains 155K scene text images and their text-erased backgrounds with detailed OCR annotations. 

The contributions can be summarized as follows:

(1) We propose a novel two-stage visual text generation method for real-world scenes called SceneVTG with a new dataset.
SceneVTG can synthesize photo-realistic scene text images, which are more suitable for real-world scenes. 

(2) In SceneVTG, we develop Text Region and Content Generator~(TRCG) that utilizes the visual reasoning ability of MLLMs to produce reasonable text regions and contents in a coarse-to-fine mechanism, and Local Visual Text Renderer~(LVTR) to enable text generation at arbitrary scale.

(3) Extensive experiments demonstrate that SceneVTG significantly outperforms rendering-based and diffusion-based methods in terms of fidelity and reasonability. Additionally, the images generated by SceneVTG provide superior utility for text detection and text recognition tasks.

\section{Related Work}
\subsection{Text Region Generation}
\label{sec:2.1}

In contrast to the text layout generation~\cite{lee2020neural,jiang2022coarse,jiang2023layoutformer++} which organizes text layouts on a blank image according to constraints, text region generation focuses on determining the optimal placement of generated text within an image.
Although text region generation is a crucial step in text synthesis, there is currently a lack of dedicated research in this field.
SynthText~\cite{gupta2016synthetic} utilizes dense depth maps, as well as color and texture segmentation to search for suitable text regions. 
VISD~\cite{zhan2018verisimilar} ensures that the suggested text regions are located in semantically consistent areas based on semantic segmentation maps and visual saliency maps. 
UnrealText~\cite{long2020unrealtext} leverages the 3D scene information of the image and probes around object meshes to find proper text regions. 
Different from the previous rule-based approach for text region generation, 
LBTS~\cite{tang2023scene} trains a segmentation network to find regions in images suitable for writing text.

However, these methods ignore the relationship between text regions and image context when generating text regions.
For example, it is unreasonable to generate text on the sky, even though the texture of the sky is consistent. 
Meanwhile, the randomly selected text contents generated by these methods are unrelated to the image context.
Inspired by the remarkable image captioning and understanding capabilities of MLLMs~\cite{liu2023visual,liu2023improved,yang2023dawn}, SceneVTG takes these issues into account by leveraging an MLLM as a text region and content generator. 

\subsection{Visual Text Generation}
Typically, visual text generation includes two types, \ie, whole image generation and sub-image generation. 
Whole image generation generally refers to generating a full image containing all visual text, and the traditional methods~\cite{gupta2016synthetic,long2020unrealtext,zhan2018verisimilar,tang2023scene} follow a three-stage process, \ie, locating suitable text regions via various feature maps, randomly choosing text content, and applying manually defined rules to render text onto the background images. 
Such rule-based rendering greatly limits the fidelity of the generated image. 
In recent years, diffusion-based visual text generation has gradually emerged. 
GlyphControl~\cite{yang2024glyphcontrol} leverages the rendered glyph image as a condition to enhance the performance of the off-the-shelf Stable-Diffusion model~\cite{Robin2022latentdiffusion}. 
TextDiffuser~\cite{chen2023textdiffuser} applies character-level segmentation masks to improve the accuracy of rendered text. 
Further, TextDiffuser-2~\cite{chen2023textdiffuser2} attempts to tame an additional language model for layout planning to generate more diverse images. 
AnyText~\cite{tuo2023anytext} first addresses multilingual visual text generation and combines multiple OCR modules to improve the performance of generated text. 
The diffusion-based visual text generation methods make the text more integrated with the image background. 
However, these methods are based on latent generation methods such as Stable-Diffusion, demonstrating limited diversity in terms of text region and text scale, exhibiting limitations in real-world scenes. 

Sub-image generation generally refers to generating localized images containing only a single text line.
In the early days, MJSynth~\cite{Jaderberg16mjsynth} rendered images through various materials and several manually defined steps. 
Later, GAN-based methods~\cite{Kang2020ganwriting,fogel2020scrabblegan, Bhunia2021handt,gan2021higan,luo2022slogan,kong2022look} emerged, which can better control the text style. 
Recently, several diffusion-based methods~\cite{yz2023ctig, Gui2023zero, Luhman2020diffusion, Nikolaidou2023WORD, Yang2023fontdiff} have proved that the diffusion model can achieve superior fidelity on sub-image generation. 
These methods usually focus on the diversity and controllability of the foreground text style, without considering the relationship between background images and foreground text, thus bringing advantages to the quality of foreground text. 
Some editing methods~\cite{omri2022blended, couairon2022diffedit, DBLP:conf/cvpr/BrooksHE23} excel in considering the relationship between the foreground and background, but they are not designed for text generation and struggle to generate high-quality text.
In this paper, we exploit the advantages of sub-image generation and consider the integration of text and background in the whole image to achieve high-fidelity image generation in real-world scenarios. 

\section{Methodology}
\begin{figure}[t]
\centering
\includegraphics[width=0.99\textwidth]{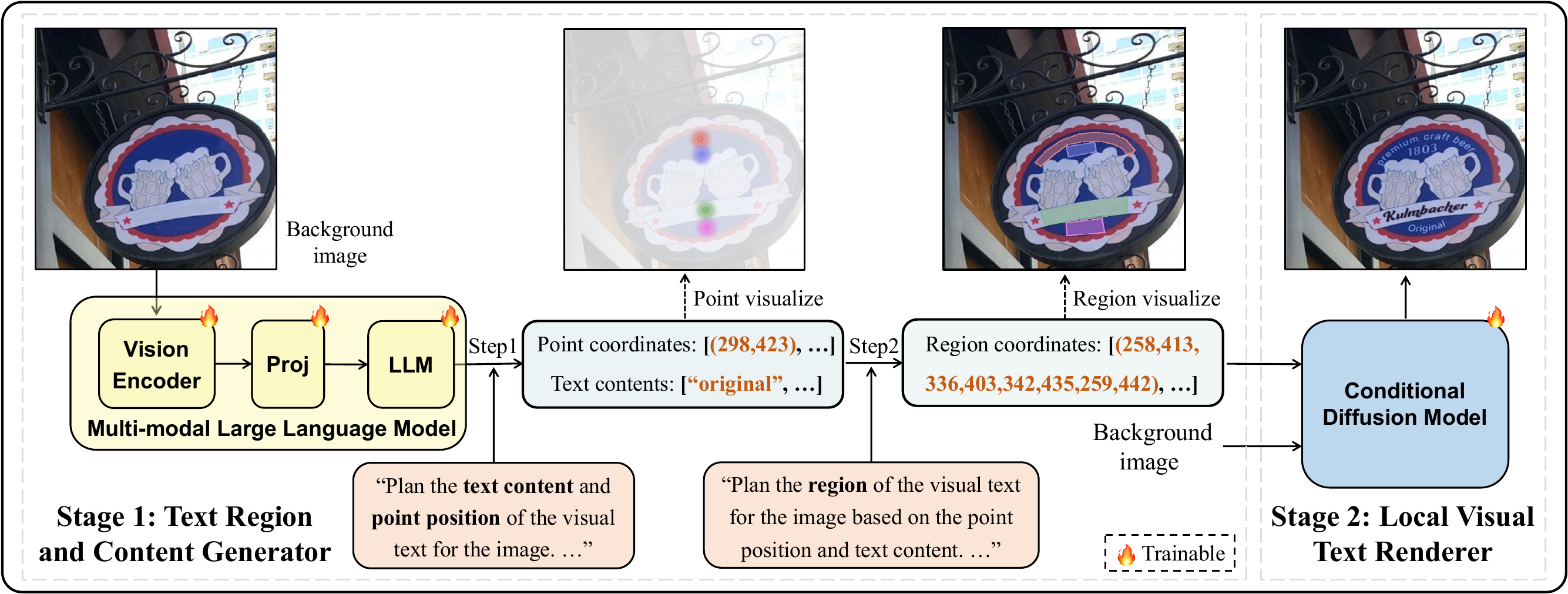}
\caption{The overall architecture of SceneVTG. Given the background image and pre-defined text prompt, SceneVTG generates text regions and contents in two steps with an MLLM and then generates visual text image with a local conditional diffusion model.} 
\label{fig:fig3}
\end{figure}

As shown in~\cref{fig:fig3}, our proposed method consists of a Text Region and Content Generator~(TRCG) and a Local Visual Text Renderer~(LVTR).
Given the background images, the TRCG generates text regions and contents in language-format.
Then, the LVTR constructs various conditions to generate local text images. 
Eventually, the resulting local text images will be fused back into the background images to obtain visual text images in the wild. 

\subsection{Text Region and Content Generator}
Recent research has demonstrated that MLLMs play a crucial role in various reasoning tasks, including reasoning detection~\cite{wei2023lenna} and reasoning segmentation~\cite{lai2023lisa}, owing to their powerful visual reasoning capabilities.
Inspired by this, we attempt to fine-tune an MLLM to recommend text region and content, aiming to achieve visual text generation with high reasonability. 
Specifically, we leverage LoRA to perform efficient fine-tuning on MLLM and enable training of the visual backbone and the projection layer. 
As demonstrated in~\cref{fig:fig3}, TRCG takes pre-defined prompts and a background image as input, and outputs text regions and contents in language-format.
Compared to previous work, our method does not require additional feature maps (\eg, dense depth, semantic segmentation, 3D scene information, \etc), making it more convenient to use.

For the pre-defined prompts, we decompose the task into two steps inspired by ReAct~\cite{yao2022react}.
In the first step, TRCG locates reasonable key points for rendering text and generates semantically reasonable text contents for each position. 
In the second step, based on the key points and text content, TRCG identifies detailed text regions. 
There are two reasons for this design: (1) It decomposes a complex task into two easier progressive tasks, which aligns with the chain-of-thought of LLMs in answering questions. (2) The second stage can refer to the character count of text contents to better determine the region size. 
Please refer to the Appendix A for detailed instructions on input prompts. 
Considering the coherence of the generated text contents and the reasonability of text region arrangements, TRCG outputs text contents in the form of text lines. 
The bounding boxes of the text lines can be converted into the bounding boxes of each word through post-processing. 
To better simulate the diverse text forms in real scenes, TRCG can generate curve bounding boxes based on Bezier curves~\cite{liu2020abcnet} with eight Bezier control points~(See more details of curved text in Appendix B). 

\begin{figure}[t]
\centering
\includegraphics[width=0.97\textwidth]{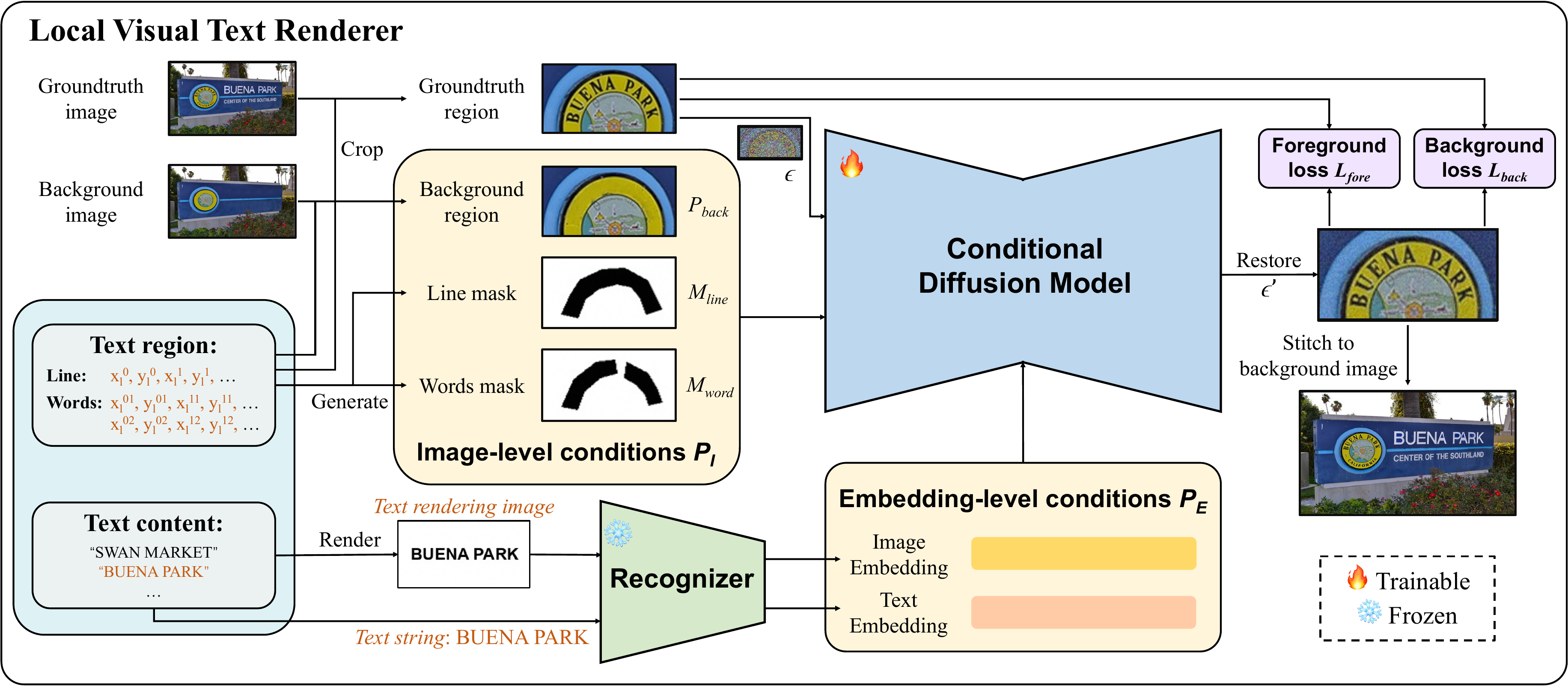}
\caption{The details pipeline of Local Visual Text Renderer. Given the TRCG outputs and background images, we construct image-level and embedding-level conditions to train the local conditional diffusion model. } 
\label{fig:fig4}
\end{figure}

\subsection{Local Visual Text Renderer}
Existing diffusion-based visual text generation methods~\cite{yang2024glyphcontrol,chen2023textdiffuser,tuo2023anytext,chen2023textdiffuser2} perform whole image generation, presenting challenges to accurately simulate the complex characteristics of text in the real world. 
Therefore, we propose a local conditional diffusion model rather than relying on latent features of the whole image, enabling text generation at arbitrary scale, and ultimately generating text images that are more consistent with real scenes. 

As presented in Fig.~\ref{fig:fig4}, given the text regions and contents generated by TRCG, we construct two major categories of conditions for the conditional diffusion model, \ie, image-level conditions~$P_I$ and embedding-level conditions $P_E$. 
Specifically, we generate a horizontal bounding rectangle based on the line region and extend the boundaries by 10\% to obtain the extended region box. The purpose of expanding the boundaries is to better restore the background information around the text during training and to retain the continuity of the original image when stitched back to the background image. 

Then, according to the extended region box, ground-truth region $P_0$ and background region $P_{back}$ are cropped, and line mask $M_{line}$ and words mask $M_{word}$ are generated. 
Among them, $P_{back}$ provides background information, while $M_{line}$ and $M_{word}$ constrain the generated text to accurately fall within the corresponding region, ensuring that the generated samples have high-quality OCR annotations to augment utility.  
Finally, $P_{back}$, $M_{line}$, and $M_{word}$ form image-level conditions $P_I$ and are input into the diffusion model together with noised groundtruth region $P_t$~(the added noise is $\epsilon \in \mathit{N}(0,1)$). 
For text content, plain text images $I_{plain}$ are rendered with a uniform font, and then fed into a parameter-frozen OCR recognition model with the text string to extract image embedding~$E_{image}$ and text embedding~$E_{text}$, respectively. 
$E_{image}$ and $E_{text}$ form embedding-level conditions $P_E$ and are input into the diffusion model together with time embedding~$t$. 

Following the usual conditional diffusion model paradigm, LVTR applies a UNet backbone $\epsilon_{\theta}$ to predict the noise~$\epsilon$ added to the $P_0$. 
The conditional diffusion loss $L_{cdm}$ can be calculated by:
\begin{align}
L_{cdm} = ||\epsilon - \epsilon_{\theta} (P_t, P_I, P_E, t)||^2_2.  
\end{align}
Additionally, two auxiliary losses, \ie, foreground loss $L_{fore}$ and background loss $L_{back}$, are proposed to improve the generation quality. 
Specifically, based on the predicted noise $\epsilon'$ and $P_t$, the predicted region $P'$ can be restored.
Then, the foreground loss is designed to focus on the semantic-level consistency of the text contents and can be described as follows:
\begin{align}
L_{fore} = ||\mathbb{F}(M_{line}\times P_0) - \mathbb{F}(M_{line}\times P')||^2_2,  
\end{align}
where $\mathbb{F}$ is the feature extractor of a pre-trained OCR recognizer.
The background loss is proposed to further maintain the pixel-level consistency of the extended background and can be formulated as: 
\begin{align}
L_{back} = ||(1-M_{line})\times P_0 - (1-M_{line})\times P'||^2_2.  
\end{align}
Finally, the overall training objective is defined as:
\begin{align}
L = L_{cdm} + \lambda_f \times L_{fore} + \lambda_b \times L_{back}, 
\end{align}
where $\lambda_f$ and $\lambda_b$ are hyperparameters and will be discussed in ablation studies.

\section{Experiments}
Currently, there is a shortage of large-scale datasets containing pairs of scene text images and text-erased background images, which are essential for training SceneVTG.
Therefore, we construct a new dataset called \textbf{SceneVTG-Erase} for training scene visual text generation and a test set \textbf{SceneVTG-benchmark}. 
Details of datasets and implementations can be found in Appendix C and D.

\begin{table}[b]
\caption{Comparison of image generation fidelity with multiple existing methods on SceneVTG-benchmark. ``B'', ``P'', ``RC'' denote background images, prompts, and specified text regions and contents, respectively.}
    \centering
   \resizebox{\textwidth}{!}
     {
    \begin{tabular}{cccccccccc}
    \toprule
    \multirow{2}{*}{Methods}    &\multicolumn{4}{c}{End-to-end Inference}  &\multicolumn{5}{c}{Local Image Generation} \\
    \cmidrule(lr){2-5}\cmidrule(lr){6-10}
     & Inputs & FID$\downarrow$ & F$\uparrow$ & LA$\uparrow$   & Inputs & FID$\downarrow$ & FID-R$\downarrow$ & F$\uparrow$ &LA$\uparrow$ \\
    \midrule    
        SynthText~\cite{gupta2016synthetic} &  B &    48.19 &       42.87 & 55.24 & B+RC & 37.36   & 46.40 & 66.07 & 40.30\\
    \midrule    
        TextDiffuser~\cite{chen2023textdiffuser} &  P+RC  & 78.35      &\textbf{53.43}  & 28.58 & - & -& -& -& -\\
        GlyphControl~\cite{yang2024glyphcontrol} &  P+RC     & 77.95     &  49.23 & 26.31 & -& -& -& -& - \\
        AnyText~\cite{tuo2023anytext} &  P+RC  & 73.07     &  44.77 & 37.17 & - & - & - & -& - \\
        TextDiffuser-2~\cite{chen2023textdiffuser2} &  P+RC  & 88.02    &  33.61 & 33.04   & B+P+RC   & 42.00    & 47.93 & 69.26 & 26.13\\
    \midrule    
        SceneVTG &  B & \textbf{26.28}& 52.03 & \textbf{75.62} & B+RC & \textbf{27.66} & \textbf{33.34} & \textbf{75.73} & \textbf{53.21} \\
      \bottomrule
\end{tabular}
     } 
    \label{tab:tab1}
\end{table}

\subsection{Experimental Results of Fidelity}
\label{exp_fidelity}
In this subsection, we compare with the previous methods~\cite{gupta2016synthetic, chen2023textdiffuser2, tuo2023anytext,chen2023textdiffuser,yang2024glyphcontrol} in terms of image fidelity, including \textit{end-to-end inference} and \textit{local image generation}. 
Similar to previous works~\cite{chen2023textdiffuser, chen2023textdiffuser2}, we use Frechet Inception Distance (FID), as well as OCR detection and recognition metrics including F-score~(F)~\cite{Dimosthenis2015ic15} and Line Accuracy~(LA)~\cite{shi2016crnn}, to evaluate the fidelity of the generated images. 
Particularly, we design a metric to evaluate the FID in specified text Region~(FID-R), which excludes the influence of background and focuses on evaluating the text generation quality. 
The experiments are evaluated on the SceneVTG-benchmark. 

\begin{figure}[t]
\centering
\includegraphics[width=0.98\textwidth]{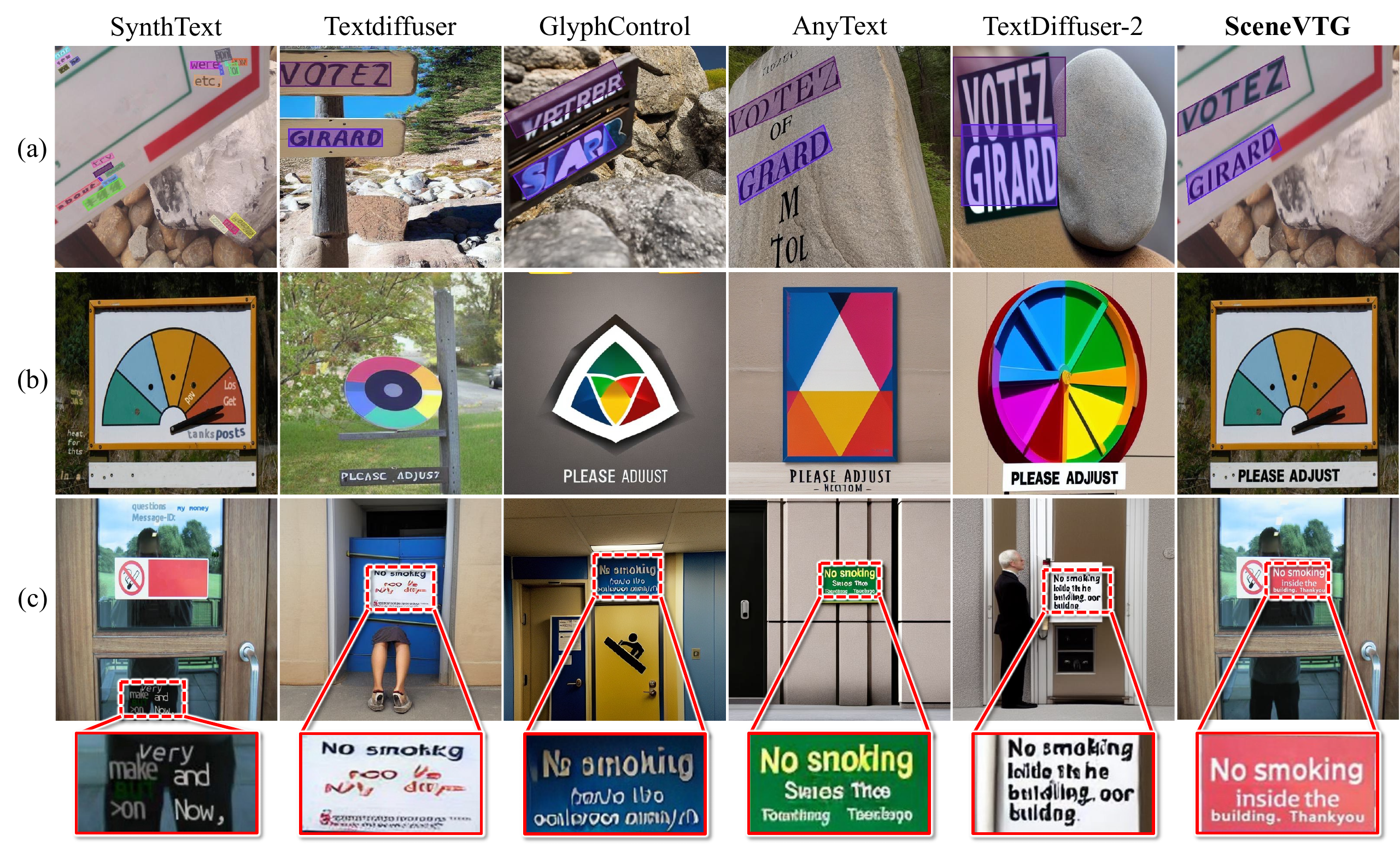}
\caption{Visualizations of end-to-end generation results compared with existing methods.
SynthText and SceneVTG automatic render visual text on background images.
TextDiffuser, GlyphControl, AnyText, and TextDiffuser-2 generate the entire images based on the same captions, text regions, and text contents.
SceneVTG can generate more accurate characters and fit better into the regions.
Zoom in for better views.} 
\label{fig:fig5}
\end{figure}

\begin{figure}[htb]
\centering
\includegraphics[width=0.98\textwidth]{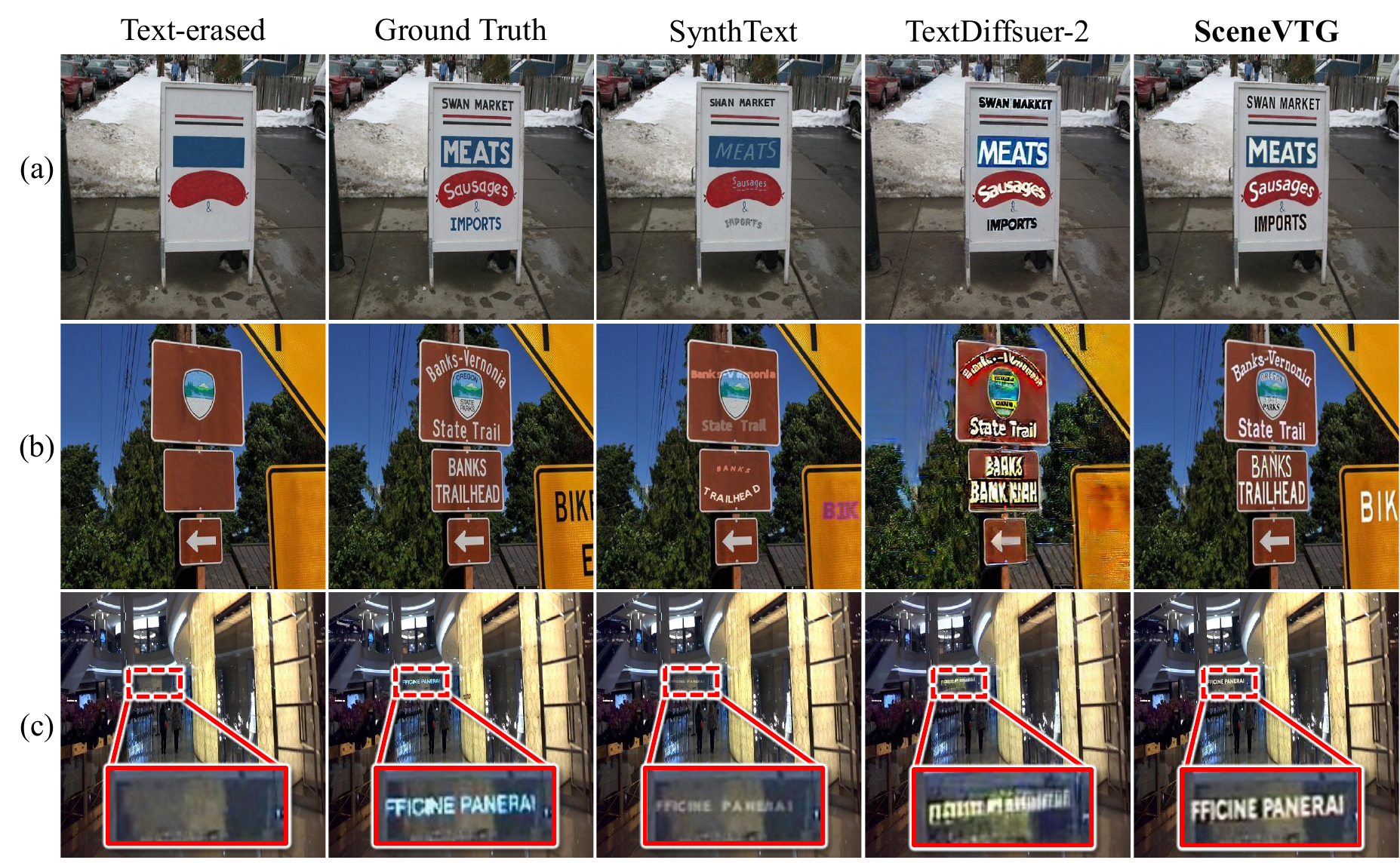}
\caption{Visualizations of local image generation results compared with existing methods. The visual text generated by SceneVTG is more harmoniously combined with the background, with no obvious artifacts. And there are no obvious errors in the text, even small text can be generated well. Zoom in for better views. } 
\label{fig:fig6}
\end{figure}

\subsubsection{End-to-end Inference.} 
Since different text image generation methods follow different paradigms, we endeavor to achieve maximal alignment in our experimental design to facilitate a fair comparison. 
Specifically, as existing diffusion-based methods cannot automatically generate visual text from text-erased background images, we obtain captions for images in the SceneVTG-benchmark by BLIP-2~\cite{li2023blip} and generate images in the text-to-image paradigm. 
Besides, text regions and contents generated by TRCG are input as text priors. 
For rendering-based methods, we only focus on SynthText~\cite{gupta2016synthetic} as other methods~\cite{zhan2018verisimilar,long2020unrealtext,tang2023scene} either lack open-source code or cannot synthesize on 2D background images. 

As shown in~\cref{tab:tab1}, SceneVTG outperforms other methods in terms of the FID evaluation metric.
Note that the diffusion-based methods show high FID since they cannot use real-world background images, showing that the backgrounds generated by these methods still exhibit a significant disparity from real-world scenes. 
For the OCR metrics, the images generated by SceneVTG are still competitive, except slightly lower than TextDiffuser on the F-score. 
The visualizations are demonstrated in~\cref{fig:fig5}. 
In~\cref{fig:fig5}~(a), the diffusion-based methods struggle to generate accurate OCR bounding boxes, and may generate additional text beyond the specified conditions due to the lax alignment in training data. In~\cref{fig:fig5}~(b) and (c), the diffusion-based methods are more likely to produce character errors, especially when rendering small-size text, while the detector may confuse these messy textures with valid text.
This may be the reason why the detection metric of TextDiffuser is high but its recognition metric is low. 
Additionally, by comparing SynthText and our method, we find the reasonability of text regions and contents makes the generated images more fidelity, and the reasonability will be discussed in detail in the next subsection. 

\subsubsection{Local Image Generation.} 
Besides the end-to-end inference to generate the whole image, we also explore the fidelity of local image generation to evaluate the quality of the generated text. 
More specifically, given a background image, the models are asked to generate specified contents in the specified regions, which is implemented as the inpainting mode in diffusion-based methods. 
FID-R is introduced to evaluate the text image quality in the corresponding regions.
The methods that lack efficient open-source code or require additional inputs to execute inpainting mode are out of the study scope.

As shown in~\cref{tab:tab1}, our SceneVTG exhibits the best performance on all evaluation metrics. \cref{fig:fig6} shows the visualizations. 
From an overall perspective, the text pixels generated by SynthText, despite being clear, can not be seamlessly integrated with the background. Note that inaccuracies in text regions owe to the limitation of open-source code.
Moreover, as presented in \cref{fig:fig6}~(a) and (b), the curved text generated by SceneVTG exhibits superior quality compared to TextDiffuser-2, with no obvious errors. Additionally, the visual integration of the text generated by SceneVTG with the background appears more seamless and natural.
Particularly, \cref{fig:fig6}~(c) shows that SceneVTG is capable of generating high-quality small-size text, a feature not present in TextDiffuser-2. 

\begin{figure}[t]
\centering
\includegraphics[width=0.8\textwidth]{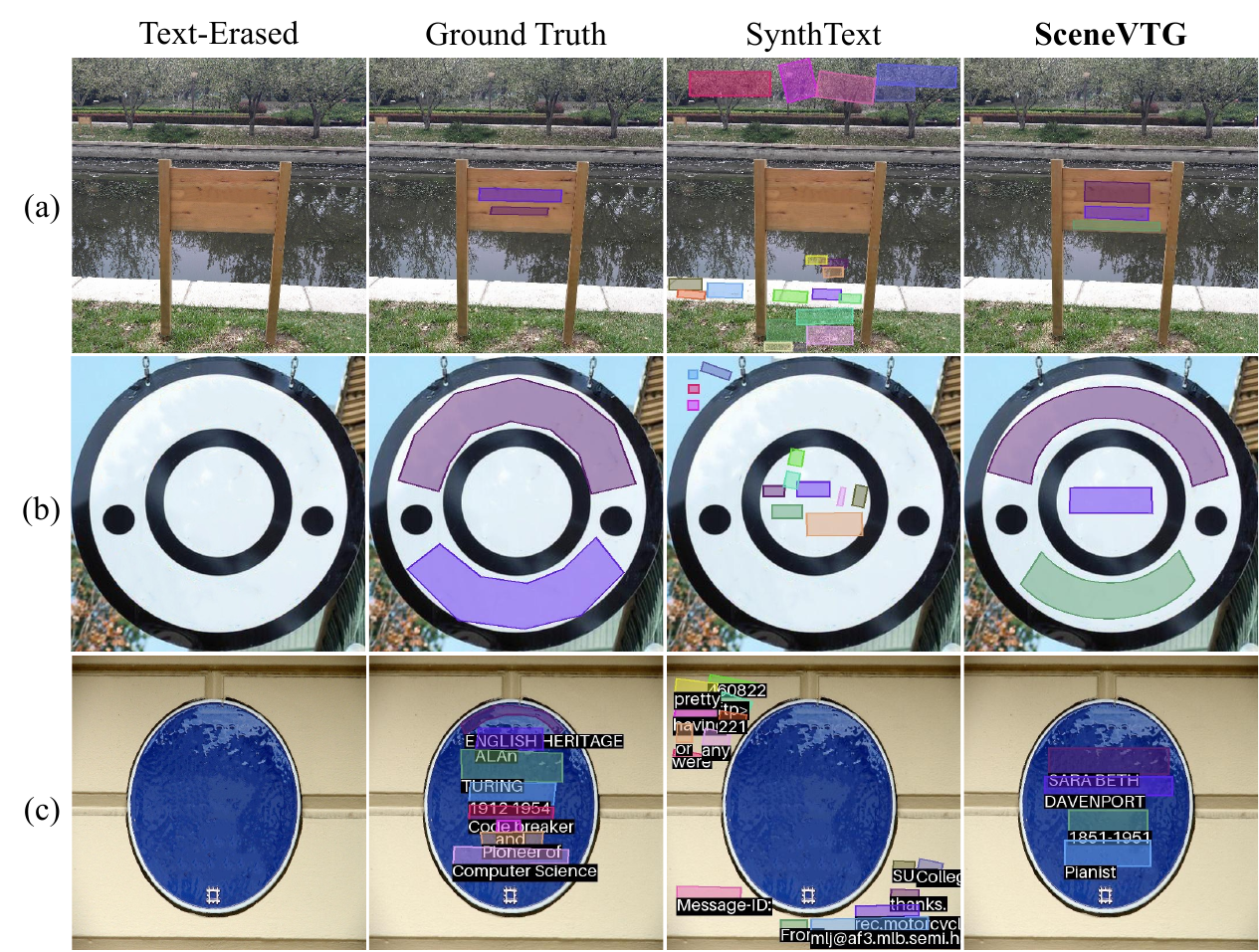}
\caption{Visualizations of text region and content generation compared to SynthText. The proposed SceneVTG exhibits reasonability, \eg, reasonable text regions in (a) and (b), and related text contents in (c), while SynthText fails to do this, \eg, text regions upon tree and river in (a), and disorganized text layout in (c). Zoom in for better views.} 
\label{fig:fig7}
\end{figure} 

\subsection{Experimental Results of Reasonability} \label{exp_reason}
\begin{table}[!b]
\caption{Comparison of reasonability with SynthText on SceneVTG-benchmark.}
    \centering
    \begin{tabular}{cccc}
    \toprule
Method     & IoU$\uparrow$  & PD-Edge$(\times1e-4)\downarrow$  & CLIPScore$\uparrow$ \\
    \midrule
    SynthText~\cite{gupta2016synthetic} & 7.31 & 1.09 & 21.74     \\
    Ours     & \textbf{33.95} & \textbf{0.68} & \textbf{22.59}      \\
    
\bottomrule
    \end{tabular} 
    
    \label{tab:reasonability}
\end{table}
In this subsection, we analyze the reasonability of the generated text regions and contents on SceneVTG-benchmark through two region-related metrics, \ie, IoU and PDEdge, and a content-related metric, \ie, CLIPScore. 
More specifically, due to the open-ended nature of the region prediction task, the predicted boxes and the ground truths cannot correspond one-to-one, we consider using the intersection and union between all predicted boxes and groundtruth boxes to calculate the IoU. 
Besides, since the text in the real world does not appear on significant boundaries in most cases, we propose a new evaluation metric PD-Edge. 
PD-Edge uses the edge detection network PiDiNet~\cite{su2021pixel} to detect edges in the image and calculate the average pixel value of the edge area covered by the generated text regions. 
CLIPScore calculates the score between the generated text contents and the background image using clip-vit-base-patch16~\cite{radford2021learning}. 

We compare the proposed TRCG with SynthText~\cite{gupta2016synthetic} since other rendering-based methods~\cite{Kang2019unsupervised,long2020unrealtext,tang2023scene} either lack open-source code or cannot synthesize on 2D background images. 
As demonstrated in \cref{tab:reasonability}, our method outperforms SynthText by a notable margin in both region-related and content-related metrics.
\cref{fig:fig7} presents some visual comparisons. 
Notably, the text regions and contents generated by our method are more reasonable when compared with SynthText. 
Concretely, in~\cref{fig:fig7}~(a), TRCG avoids generating text in unreasonable regions while SynthText generates text on the tree and river. 
In~\cref{fig:fig7}~(b), TRCG generates curved text regions to adapt to the annular areas in the background. 
In~\cref{fig:fig7}~(c), TRCG arranges multiple text regions within a holistic layout, resulting in more reasonable arrangements. Conversely, SynthText fails to do this, leading to arrangements that significantly differ from real scenes. 
Additionally, the text contents generated by SynthText are random, while those generated by our model have stronger correlations with each other and the image context. 

\subsection{Experimental Results of Utility}
\begin{table}[!b]
 \caption{
 Comparison of the effects of training text detectors and recognizers using synthetic data from different methods and real data~\cite{Baek2021what}.
 ``Regular'', ``Irregular'', and ``Avg'' denotes the average of IIIT, SVT, and IC13, the average of IC15, SVTP, and CUTE, and the average of all datasets, respectively. $\dagger$ means data copied from \cite{tang2023scene}.}
    \centering
   \resizebox{\textwidth}{!}
     {
    \begin{tabular}{ccccccccccc}
    \toprule
    \multirow{2}{*}{Methods}    &\multicolumn{5}{c}{Detection Results~(F$\uparrow$)}  &\multicolumn{5}{c}{Recognition Result~(LA$\uparrow$)} \\
    \cmidrule(lr){2-6}\cmidrule(lr){7-11}
     & Detector & Size & IC13 & IC15 & MLT17 & Recognizer & Size & Regular & Irregular & Avg \\
    \midrule    
        SynthText~\cite{gupta2016synthetic} &  EAST$\dagger$ &  10K &    72.38&      56.65&      47.17   &  CRNN &  30K  & 48.55 & 26.08 & 39.74 \\
        VISD~\cite{zhan2018verisimilar} & EAST$\dagger$  &   10K&    75.19&     65.49&     50.21 &  CRNN &  30K  & 38.02 & 26.88 & 33.65\\
        UnrealText~\cite{long2020unrealtext} &  EAST$\dagger$  &   10K &    73.73&     61.80&    47.62&  CRNN &  30K  & 32.06 & 19.85 & 27.28\\
        LBTS~\cite{tang2023scene} &  EAST$\dagger$ &  10K &    68.83&     52.69&     40.59 & -  & -  & - & - & -\\
    \midrule    
        TextDiffuser~\cite{chen2023textdiffuser} &  EAST  &   10K &    52.50&     34.08&     29.06 &  CRNN &  30K  & 30.08 & 8.79 & 21.74 \\
        GlyphControl~\cite{yang2024glyphcontrol} &  EAST &   10K &    35.04&    40.64&     24.37&  CRNN &  30K  & 40.80 & 13.21 & 29.99\\
        AnyText~\cite{tuo2023anytext} &  EAST &   10K &  48.05 &    42.45&    30.75 & CRNN &  30K  & 49.80 & 19.62 & 37.97 \\
        TextDiffuser-2~\cite{chen2023textdiffuser2} &  EAST  &  10K &    21.47&      10.79&     12.23&  CRNN &  30K  & 40.26 & 10.48 & 28.59\\
    \midrule    
        Real data & EAST$\dagger$ & - & \underline{76.34} & \underline{79.15} & \underline{56.09} & CRNN & 30K & \underline{57.97} & \underline{37.66} & \underline{50.01} \\
        SceneVTG &  EAST &  10K  &    \textbf{75.36}&     \textbf{66.31}&     \textbf{53.90} & CRNN &  30K  & \textbf{54.97} & \textbf{35.50} & \textbf{47.34} \\
      \bottomrule
    \end{tabular} 
     } 
    \label{tab:EAST}
\end{table}

In this subsection, we explore the utility of our method by evaluating the performance of generated samples on training OCR tasks, \ie, text detection and text recognition tasks~\cite{zhu2016scene, Long2018SceneTD, Yao2012DetectingTO, shi2016crnn, liao2022real, iccv2023lister, mgpstrv2, DBLP:journals/pr/YangYLZB24}. 
The detailed settings and implementations are the same as previous works~\cite{tang2023scene,yz2023ctig} and are provided in Appendix E. 
The results are presented in Tab.~\ref{tab:EAST}.
An experiment trained with the same amount of real data is conducted for reference. 
It can be found that the rendering-based methods have better detection performance than the diffusion-based methods. This is because the diffusion-based methods usually cannot obtain accurate and uniform-granularity detection boxes, as shown in Fig.~\ref{fig:fig5}.
For the recognition results, the diffusion-based methods are slightly better than the rendering-based methods on the \textit{Regular} dataset, while are much worse on the \textit{Irregular} dataset.
This may be because the samples generated by diffusion-based methods are generally of more regular shapes, as shown in~\cref{fig:fig1}. 
Finally, training OCR models with samples generated by our method can achieve the best performance on all evaluation sets, which suggests the proposed SceneVTG is able to produce valid image samples that can better simulate the real world. 
Particularly, our method improves a large margin on \textit{Irregular} dataset, which is inseparable from the ability of SceneVTG to synthesize photo-realistic curved visual text.

\subsection{Ablation Study}
\begin{table}[!b]
\caption{Ablation study for TRCG. We explore the role of fine-tuning the visual encoder~(denoted as FT VisEnc) and two steps generation~(denote as Two steps).}
    \centering
    \begin{tabular}{cccccc}
    \toprule
Exp. &FT VisEnc & Two steps    & IoU$\uparrow$  & PD-Edge($\times$1e-4)$\downarrow$  & CLIPScore$\uparrow$ \\
    \midrule
    1  &  \checkmark &  \checkmark   & \textbf{33.95} & \textbf{0.68} & \textbf{22.59}      \\
    2 &   &  \checkmark  & 22.06 & 4.80 & 22.56     \\
    3 &  \checkmark &    & 7.97 & 3.92 & 22.26 \\
    
\bottomrule
    \end{tabular} 
    
    \label{tab:ablation_stage1}
\end{table}
\begin{table}[htbp]
  \centering
  \caption{Ablation study for LVTR. We investigate the role of region masks, auxiliary losses, and optimal values of hyperparameters.}
    \begin{tabular}{ccccccccc}
    \toprule
    Exp.   & Region masks     & $L_{fore}$    & $\lambda_f$    & $L_{back}$    & $\lambda_b$    & FID-R$\downarrow$  & F$\uparrow$     & LA$\uparrow$ \\
    \midrule
    1     & \checkmark     & \checkmark     & 0.01   & \checkmark     & 1.0    & 33.34 & \textbf{75.73} & \textbf{53.21} \\
    2     &        & \checkmark     & 0.01  & \checkmark     & 1.0    & 40.75 & 68.17 & 47.64 \\
    3     & \checkmark       &       &    & \checkmark     & 1.0      & 39.66 & 70.41 & 49.15 \\
    4     & \checkmark      & \checkmark     & 0.03   & \checkmark     & 1.0   & 38.01 & 73.77 & 49.79 \\
    5     & \checkmark     & \checkmark     & 0.003    & \checkmark     & 1.0   & \textbf{32.75} & 75.04 & 53.01 \\    
    6     & \checkmark     & \checkmark     & 0.01   &       &      & 36.96 & 75.71 & 52.63 \\
    7     & \checkmark    & \checkmark     & 0.01     & \checkmark     & 0.3  & 37.32 & 74.88 & 50.99 \\

    \bottomrule
    \end{tabular}%
  \label{ablation_stage2}%
\end{table}%

We conduct ablation experiments on the SceneVTG-benchmark in this subsection for each stage in SceneVTG.
\cref{tab:ablation_stage1} shows the result for TRCG and reveals the following conclusions:
$({\romannumeral1})$\textit{Role of fine-tuning the visual encoder}. 
By comparing Exp.1 and Exp.2, when we disable fine-tuning the vision encoder, both the performance of text region generation and text content generation in the model will decline. 
Among them, the PD-Edge shows the largest change. 
We hypothesize that this is because the pre-trained vision encoder is not sensitive to the edge regions in the image, making it easy to cover the edge regions when generating text regions. 
$({\romannumeral2})$\textit{Role of two steps generation}. 
By comparing Exp.1 and Exp.3, when we try to generate text regions and text contents without taking two steps, the performance of the model shows a notable decline, with the text content generation ability degraded to a level similar to random generation. 
This phenomenon indicates that it is necessary to decompose the tasks of text region generation and text content generation into two steps.

Regarding LVTR, We further investigate the role of region masks~(\ie, $M_{line}$ and $M_{word}$), auxiliary losses~(\ie, $L_{fore}$ and $L_{back}$) and hyperparameters~(\ie, $\lambda_f$ and $\lambda_b$).
Note that $P_{back}$, $E_{image}$ and $E_{text}$ are basic conditions and are not discussed here, since it is naive to understand that without them the backgrounds and text contents will be unconditional, leading to poor fidelity. 
Tab.~\ref{ablation_stage2} shows the experiment results. 
$({\romannumeral1})$\textit{Role of region masks}. 
By comparing Exp.1 and Exp.2, when we disable region masks, a series of fidelity metrics decrease, especially the accuracy of boxes and text. 
This is naive since without the conditional control of the region mask, the region to generate text will become unconstrained, resulting in inaccurate text pixels, ultimately affecting the fidelity.
$({\romannumeral2})$\textit{Role of $L_{fore}$ and choice of $\lambda_f$}. 
By comparing Exp.1 and Exp.3, we conducted experiments without using  $L_{fore}$ and found that the authenticity of the image will be significantly reduced, which shows that additional supervision of foreground text is very necessary, especially for the accuracy of text. 
Furthermore, in Exp.3, Exp.4, and Exp.5, we explore the optimal value of $\lambda_f$ and use it as our default setting. 
$({\romannumeral3})$\textit{Role of $L_{back}$ and choice of $\lambda_b$}. 
By comparing Exp.1 and Exp.6, when we disable $L_{back}$, the FID of the generated images decreases somewhat but not significantly. 
We suspect that this is because FID evaluation may not be obvious for pixel-level differences, and the loss of the diffusion model itself also assists in supervising the consistency of the background. 
However, we have empirically found that the LVTR model converges faster when adding $L_{back}$, so we still recommend utilizing $L_{back}$ during training. 

\section{Conclusion and Limitations}
In this paper, we delve into the problem of visual text image generation in the wild.
To address this challenging problem, we propose a novel two-stage method called SceneVTG, which consists of a Text Region and Content Generator (TRCG) and a Local Visual Text Renderer (LVTR), to synthesize photo-realistic scene text images.
TRCG leverages the visual reasoning ability of MLLMs to generate reasonable text regions and contents for the background images. 
LVTR utilizes a local conditional diffusion model to enable text generation at arbitrary scales with high fidelity. 
To train our models, we contribute a new dataset SceneVTG-Erase, which contains 155K scene text images and their text-erased backgrounds with detailed OCR annotations. 
Extensive experiments verified the fidelity, reasonability, and utility of our proposed SceneVTG. 
Limited by the type and annotation of the current training set, the text colors and fonts in the images generated by SceneVTG are relatively single and uncontrollable. 
Moving forward, these text priors can be recommended to achieve better fidelity and reasonability, and controllability can be improved to meet individual requirements.
Additionally, multi-language support (\eg, Chinese) and an end-to-end pipeline are also valuable to research in the future. 

\newpage
\section*{Acknowledgements} 
This work was supported by the National Science and Technology Major Project under Grant No. 2023YFF0905400 and the Alibaba Innovative Research (AIR) program. 

\bibliographystyle{splncs04}
\bibliography{egbib}

\clearpage
\appendix




\section{Detailed instructions on input prompts in TRCG}
\subsection{Step one: key points and text contents generation}

\begin{table}[ht]\centering

\begin{minipage}{0.99\columnwidth}    \centering
\begin{tcolorbox} 
    \centering
    \small
    \begin{tabular}{p{0.99\columnwidth}}

\begin{minipage}{0.99\columnwidth}

``conversations'': [\{``from'': ``human'', ``value'': 
``Given a background image that will be written with text, plan the text and location of the visual text for the image. \
The planned locations are represented by the coordinates of the points, and they should be located in suitable areas of the image for writing text. \
The size of the image is 512*512. Therefore, none of the properties of the positions should exceed 512. \
The planned texts should be related to each other and fit to appear in the location represented by the corresponding point. \
The location and text of the planned visual text need to be represented in JSON format.''\}]
\end{minipage}
    \end{tabular}
\end{tcolorbox}
    
\caption{Prompt for step one. The model will output key points and text contents.}
    \label{tab:prompt_step1}
\end{minipage}
\end{table}

\subsection{Step two: text regions generation}

\begin{table}[ht]\centering

\begin{minipage}{0.99\columnwidth}    \centering
\begin{tcolorbox} 
    \centering
    \small
    \begin{tabular}{p{0.99\columnwidth}}

\begin{minipage}{0.99\columnwidth}

``conversations'': [\{``from'': ``human'', ``value'': 
``Based on the point and text planned above, plan the layout of the visual text for the image. \
Point guide where the layout should be and the planned layout should be located near the point. \
Layouts are represented in the form of Bezier curve control point coordinates, representing an area on an image suitable for writing visual text. \
Each box consists of eight vertices, starting at the top left corner in counterclockwise order. \
The appropriate layouts should not overlap each other. \
The aspect ratio of the layout boxes needs to consider the number of characters in the texts, the more characters, the larger the aspect ratio. \
The layout and text of the planned visual text need to be represented in JSON format too.''\}]
\end{minipage}
    \end{tabular}
\end{tcolorbox}
    
\caption{Prompt for step two. The model will output text regions and text contents.}
    \label{tab:prompt_step2}
\end{minipage}
\end{table}

\section{Details for curved text region generation}
Following ABCNet~\cite{liu2020abcnet}, the curved texts are represented by the Bezier curve. 
It employs the Bernstein Polynomials to parameter curve text $c(t)$ as:
\begin{align}
c(t)=\sum_{i=0}^{n}b_{i}B_{i,n}(t), 0<=t<=1, 
\label{eq:bezier}
\end{align}
where $n$ represents the degree of the Bezier curve. The $i$-th control point is represented by $b_{i}$, and the Bernstein basis polynomials are denoted as $B_{i,n}(t)$. 
Referring to the conclusion of ABCNet, we use cubic Bezier curve~($n=3$), \ie, eight Bezier control points, to represent all curve texts. 
Since we find that the pre-training knowledge of the MLLM contains the knowledge of Bezier curves, and can understand Bezier curves and their corresponding control points, there are no additional modules for the generation of Bezier curves.

\section{Details for SceneVTG-Erase datasets}
\subsection{Construction process}
\begin{table}[t]
\caption{Details of the sources of the SceneVTG-Erase.
``Introduce'' denotes the subsets of each dataset introduced into SceneVTG-Erase.
$\checkmark$ under ``Line annotation'' indicates that the original dataset contains line-level annotations. We have manually supplement line-level annotations for other datasets.
}
    \centering
   \resizebox{0.99\textwidth}{!}
     {
    \begin{tabular}{lccccc}
    \toprule
    Dataset & Introduce & Line annotation & Image count & Line count & Ratio\\
    \midrule    
        SCUT-EnsText~\cite{liu2020erasenet} & train &  & 2,176  & 12,632 \\
        CTW-1500~\cite{yuliang2017detecting} & train & \checkmark & 1,000 & 7,696 \\
        ICDAR-2013~\cite{karatzas2013icdar} & train &  & 229 & 526 \\
        ICDAR-2015~\cite{karatzas2015icdar} & train & & 1000 & 8,734 \\
        TextOCR~\cite{singh2021textocr} & train &  & 21,778 & 723,004 \\
        TotalText~\cite{ch2020total} & train &  & 1,255 & 5,785 \\
        HierText~\cite{long2023icdar} & train\&val &  & 10,005 & 596,112 \\
        UberText~\cite{zhang2017uber} & train\&val\&test & \checkmark & 117,969 & 571,534 \\
    \midrule    
        Total count & & & 155,412 &  1,926,023 \\
    \midrule    
        Language \\
        ~~~- English & & & & 1,862,448 & 96.70\% \\
        ~~~- Chinese & & & & 686 & 0.03\% \\
        ~~~- Others & & & & 62,889 & 3.27\% \\
        Scales~(Height) \\
        ~~~- Pixel~$\leq$~8 & & & & 274,767 & 14.26\%\\
        ~~~- 9~$\leq$~Pixel~$\leq$~64 & & & & 1,479,567 & 76.82\%\\
        ~~~- 65~$\leq$~Pixel & & & & 171,689 & 8.92\%\\
        Content \\
        ~~~- Unclear text & & & & 401,723 & 20.86\%\\
        ~~~- 65~$\leq$~Chars & & &  & 9,488 & 0.49\%\\
      \bottomrule
    \end{tabular} 
     } 
    
    \label{tab:app_tab1}
\end{table}
We construct \textbf{SceneVTG-Erase} by collecting data from two distinct sources. 
On the one hand, we incorporate the public scene text removal dataset SCUT-EnsText~\cite{liu2020erasenet} and use its training set. 
On the other hand, we collect various public scene text datasets, including the CTW-1500~\cite{yuliang2017detecting}, ICDAR-2013~\cite{karatzas2013icdar}, ICDAR-2015~\cite{karatzas2015icdar}, TextOCR~\cite{singh2021textocr}, TotalText~\cite{ch2020total}, HierText~\cite{long2023icdar}, and UberText~\cite{zhang2017uber}, and apply SOTA scene text removal method CTRNet~\cite{liu2022don} on these datasets to generate text-erased background. 
After erasing, we use DiffBIR~\cite{lin2023diffbir} to perform image restoration to smooth the artificial traces in the erased area. 
To maintain the unity of text style in the same line, the basic unit of LVTR training is text line. 
Since most of the subsets that make up SceneVTG-Erase do not have text line annotations, we complete them through manual annotation. 
Finally, SceneVTG-Erase contains about \textbf{155K} whole image pairs and \textbf{1.92M} sub-image pairs. 
\cref{tab:app_tab1} illustrates the detailed statistics of the proposed dataset. 
Specifically, the language of the data in SceneVTG-Erase is mainly English (about 96.7\%). Our work focuses on English scenes and excludes data in other languages. 
The scale of the data is mainly concentrated in medium resolution with text line height ranging from 8 to 64 pixels. 
Additionally, it also includes a numerous text with uncertain characters.
Considering the high quality of SCUT-EnsText, which was manually annotated with PhotoShop, we chose its test set as our evaluation set \textbf{SceneVTG-benchmark}, which has 646 image pairs. 
We hope the proposed SceneVTG-Erase can contribute to more scene text research in the community.

\subsection{Visualization}
\begin{figure}[t]
\centering
\includegraphics[width=\textwidth]{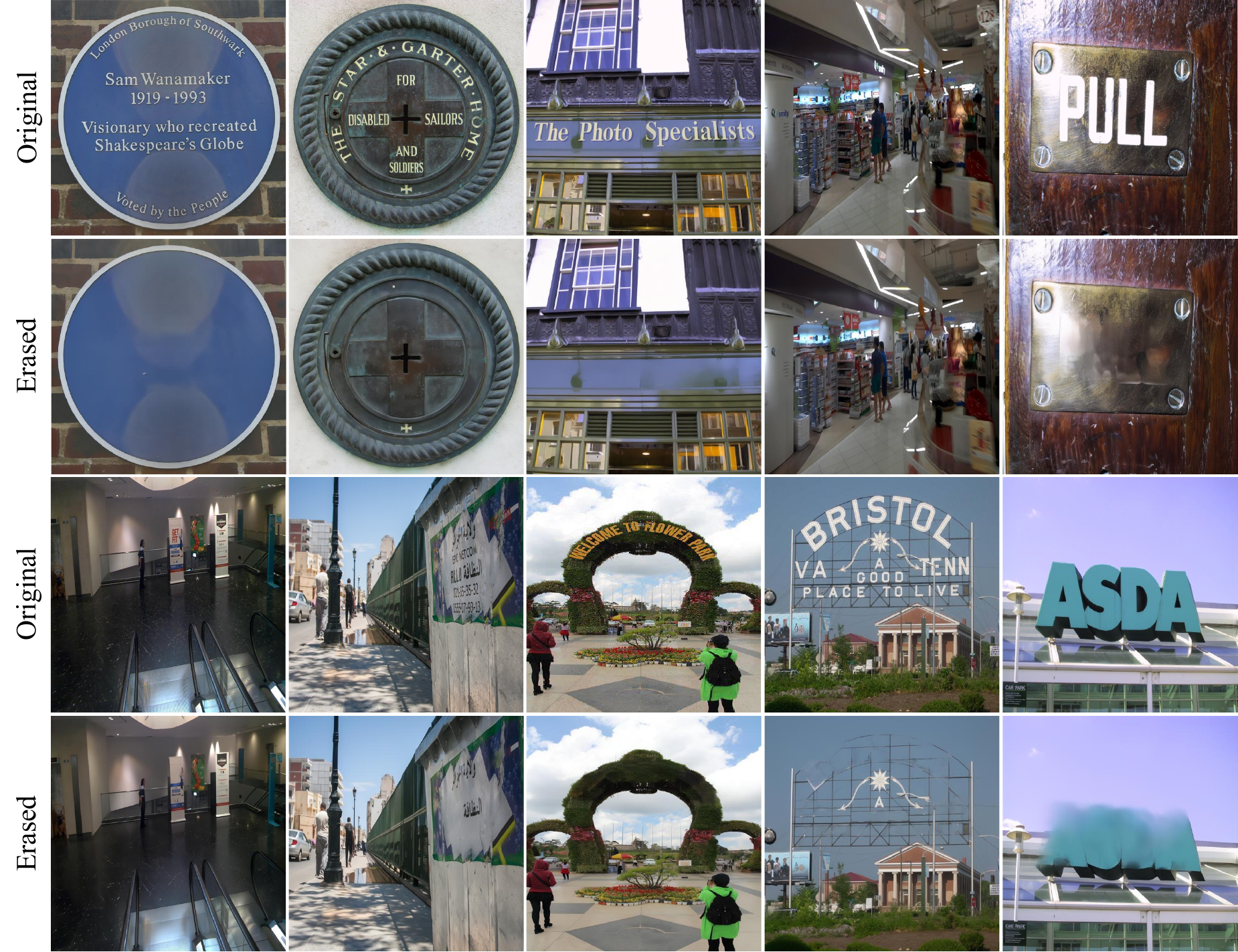}
\caption{Examples of training images including original images and erased images. 
The erasure methods work well for both simple and complex backgrounds, except for the last image which has no clear carrier.
Zoom in for better views.} 
\label{fig:app_fig1}
\end{figure}

\cref{fig:app_fig1} shows some examples of the training images before and after erasure.
It can be seen that our processing method can achieve great erasure performance for most images, especially when there are clear text carriers. 

\subsection{Filtering process}
TRCG and LVTR are trained independently and obtain their respective training subsets through the following filtering rules.
For \textbf{TRCG}, each original image undergoes the following filtering rules:
\begin{itemize}
    \item Number of text lines in the image should not exceed 12.
    \item The Height and width of the image should not be larger than 2048.
    \item For the 4K resolution images in the UberText dataset, we crop them into four 2048*2048 images. 
\end{itemize}
Afterward, the remaining images undergo a filtering process where each text line is subjected to the following rules: 
\begin{itemize}
    \item Height of the text should not be less than 8 pixels.
    \item The number of characters in each text line does not exceed 64.
    \item There cannot be any uncertain character in the text content annotation. 
    \item The IOU between the annotation represented by the Bezier curve box and the original polygon box annotation is greater than 0.8.
\end{itemize}
After strict filtering, about 44K images with 117K text lines were used for TRCG. For \textbf{LVTR}, all text lines are subjected to the following filtering rules: 
\begin{itemize}
    \item Height of the text should not be less than 8 pixels.
    \item The number of characters in each text line does not exceed 64.
    \item There cannot be any uncertain character in the text content annotation.
\end{itemize}
Since the training unit of LVTR is text line, there is no need to consider whether all text lines in the whole image are suitable for training. 
Finally, a total of 1.07M text lines were selected for training LVTR. 
TD language and text regions.

\section{Implementation Details}
Since TRCG and LVTR use different structures to optimize different objectives, they are trained independently but perform end-to-end inference. 
For \textbf{TRCG}, we fine-tune the pre-trained LLaVA-v1.5-7B~\cite{liu2023visual} model on the SceneVTG-Erase dataset. 
In addition to fine-tuning the projection layer and the large language model, we also enable the fine-tuning of the visual encoder.
The learning rate for all three modules is $2e-5$, with the large language model being fine-tuned using LoRA~\cite{hu2021lora}.
The LoRA training was conducted with rank $r=128$ and $\alpha=256$.
We fine-tune for a total of 16 epochs with a batch size of 16, which requires one day of training using 4 Tesla A100 GPUs. 
\textbf{LVTR} is implemented based on DALLE-2~\cite{Aditya2022unclip} without pre-trained weight. 
The pre-trained text recognizer uses CRNN architecture~\cite{shi2016crnn}. 
LVTR is trained on the SceneVTG-Erase dataset for 100 epochs using 4 Tesla A100 GPUs.
We use the AdamW optimizer with a learning rate of $1e-5$ and a batch size of 96.
The local region size is set to be 64~$\times$~512 and the drop ratio of $E_{image}$ and $E_{text}$ are set to 50\% and 20\% respectively. 
In the first 50 epochs, we enable the $L_{back}$ with a weight coefficient of $\lambda_b$ = 1.0. 
In the last 50 epochs, we disable the $L_{back}$ and enable the $L_{fore}$ with a weight coefficient of $\lambda_f$ = 0.01. 

\section{Detailed experiments setting of utility}

\subsection{Detection}
\label{detection}
Following the settings in LBTS~\cite{tang2023scene}, We select EAST~\cite{zhou2017east} as the baseline text detector and use ResNet-50 as the backbone. 
All the models are trained on two Tesla V100 GPUs with a batch size of 24 and a total of 200,000 steps. 
The performance metric is F-score~(F), calculated under the ICDAR2015 evaluation protocol~\cite{Dimosthenis2015ic15} overall evaluation datasets. 

For fair comparisons, all methods use \textbf{10K} synthesis samples, and to achieve the best performance of each method, we used the datasets they proposed. 
More, specifically, for \textbf{SynthText}~\cite{gupta2016synthetic}, \textbf{VISD}~\cite{zhan2018verisimilar} and \textbf{UnrealText}~\cite{long2020unrealtext}, the training sets are constructed from their open source synthetic data. 
\textbf{LBTS} do not open source the synthesized images, so we use the results reported in their paper. 
\textbf{Textdiffuser}~\cite{chen2023textdiffuser} proposed MARIO-Eval benchmark and we use it to construct the training set for Textdiffuser and \textbf{Textdiffuser-2}~\cite{chen2023textdiffuser2}. 
The training set for \textbf{GlyphControl}~\cite{yang2024glyphcontrol} are selected from LAION-Glyph-1M they proposed. 
\textbf{AnyText}~\cite{tuo2023anytext} proposed AnyText-benchmark and we use it to construct its training set. 
For our \textbf{SceneVTG}, since the SceneVTG-benchmark only has 813 samples, we conduct the same processing on COCO-Text~\cite{veit2016coco} as we did in constructing SceneVTG-Erase to make up for the insufficient background images.

The evaluation datasets are \textbf{IC13}~\cite{karatzas2013icdar}, \textbf{IC15}~\cite{karatzas2015icdar} and \textbf{MLT17}~\cite{nayef2017icdar2017}. 
IC13 includes 233 test images in English.
IC15 comprises 500 test images in English. 
MLT17 contains 1,800 test images from nine different languages. 

\subsection{Recognition}
Following the settings of previous work~\cite{yz2023ctig}, we select CRNN~\cite{shi2016crnn} as the baseline text recognizer. 
All the models are trained on one Tesla V100 GPU with a batch size of 64 and a total of 100,000 steps. 
The performance metric is Line Accuracy~(LA), which denotes the proportion of test samples whose recognition results completely match the groundtruths. 

All methods use the entire images used for detection to crop the corresponding sub-images for recognition. 
Since the amounts of sub-images in the generated images of different methods are inconsistent, we all use \textbf{30K} randomly selected from them to make a fair comparison. 
Besides, the real data are randomly selected from Real-L~\cite{Baek2021what}.
Experiments are conducted on \textbf{IIIT}~\cite{Mishra2012iiit5k}, \textbf{SVT}~\cite{Wang2011svt}, \textbf{IC13}~\cite{Dimosthenis2013ic13}, \textbf{IC15}~\cite{Dimosthenis2015ic15}, \textbf{SVTP}~\cite{Trung2013svtp} and \textbf{CUTE}~\cite{Anhar2014cute}, which contain 3,000, 647, 1,015, 2,077, 645, and 288 test images, respectively. 
According to the difficulty and geometric layout of the texts, these benchmark datasets are divided into ``regular'' or ``irregular'' datasets~\cite{shi2016robust,yang2017learning}. 
Specifically, IIIT, SVT, and IC13 are regular datasets dominated by text images with horizontally laid-out characters. IC15, SVTP, and CUTE are irregular datasets that typically contain harder cases, such as curved and arbitrarily rotated or distorted texts. 

For the fairness of the experiment, it should be noted that the training set SceneVTG-Erase has eliminated the test data that may be involved in this part.

\section{More experiments}
\subsection{Performance across different scales}
We provide the performance of LVTR generation across different scales of text regions to further quantify SceneVTG's ability to generate small-scale text. 
Specifically, we divide the samples into small-scale (pix $\leq$ 16), medium-scale (17 $\leq$ pix $\leq$ 32), and high-scale (33 $\leq$ pix) according to their text line height, and evaluate the performance of LA in local image generation mode at different scales. 
The results are presented in \cref{tab:scales}. Our SceneVTG outperforms Textdiffuser-2 in generating text at smaller scales. 
\begin{table}[htbp]
\caption{The LA score of local image generation mode across different scales.}
    \centering
    \begin{tabular}{ccccccccc}
    \toprule
Method   &  & Pixel~$\leq$~16  & & 17~$\leq$~Pixel~$\leq$~32 & & 33~$\leq$~Pixel & &Avg \\
    \midrule
    Textdiffuser2~\cite{chen2023textdiffuser2} & & 0.92 & & 14.08  & & 37.35  &  &26.13    \\
    Ours    &  & \textbf{45.16} & & \textbf{63.51} & & \textbf{50.36}  & & \textbf{53.21}    \\
    
\bottomrule
    \end{tabular} 
    \label{tab:scales}
\end{table}

\subsection{User study}
We conducted the user study on local image generation mode and reasonability task since they use the same background image in different methods, which excludes the influence of background image. 
For the user study of reasonability, we randomly select 30 groups of images (each containing SynthText, ours, and GT) and let 20 users rate the reasonability of the images from 1 to 3. The results are 1.01, \textbf{2.37}, and \underline{2.62}, respectively. 
For the user study of local image generation, we randomly select 30 groups of images (each containing SynthText, Textdiffuser-2, ours, and GT) and let 20 users rate the fidelity of the images from 1 to 4. The results are 1.74, 1.77, \textbf{2.84} and \underline{3.63}, respectively. 
These results demonstrate that the proposed SceneVTG surpasses other methods and is closer to the groundtruth score. 

\subsection{Diversity}
\begin{figure}[ht]
\centering
\includegraphics[width=0.98\textwidth]{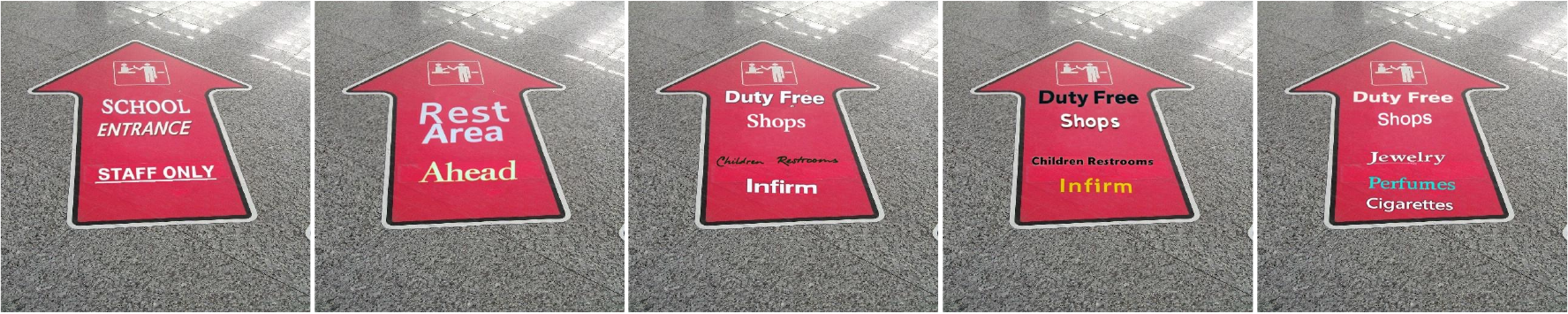}
\caption{Diversity of generated samples.} 
\label{fig:app_div}
\end{figure}

The diversity of the generated visual text images is shown in \cref{fig:app_div}. The images are generated on the same background image, with diverse text appearance, region, and content.

\section{More visualizations of generation samples}
\subsection{End-to-end generation}
\begin{figure}[ht]
\centering
\includegraphics[width=0.98\textwidth]{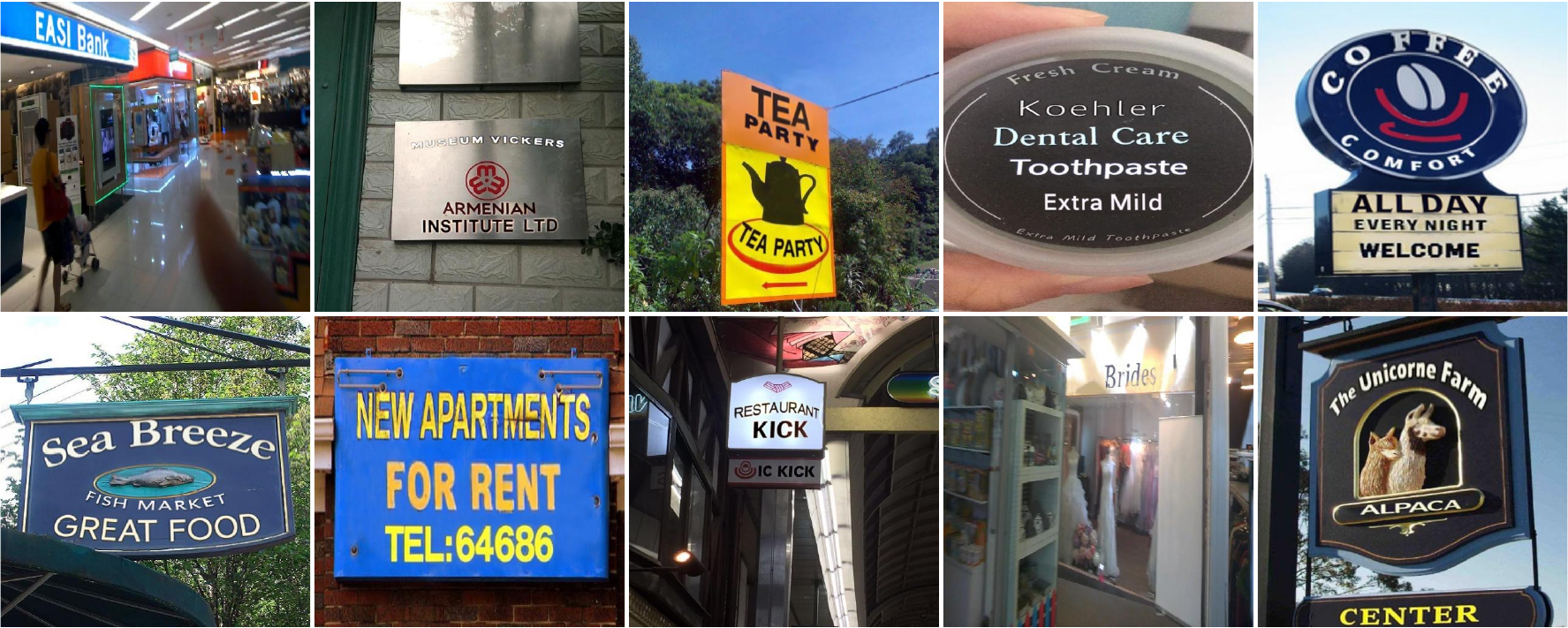}
\caption{More end-to-end generation samples on SceneVTG-benchmark.} 
\label{fig:app_fig2}
\end{figure}

\cref{fig:app_fig2} presents more end-to-end generation~(\textit{end-to-end inference} mode in Sec.~5.2 of the main text) samples on SceneVTG-benchmark. 

\subsection{LVTR generation}
\begin{figure}[ht]
\centering
\includegraphics[width=0.98\textwidth]{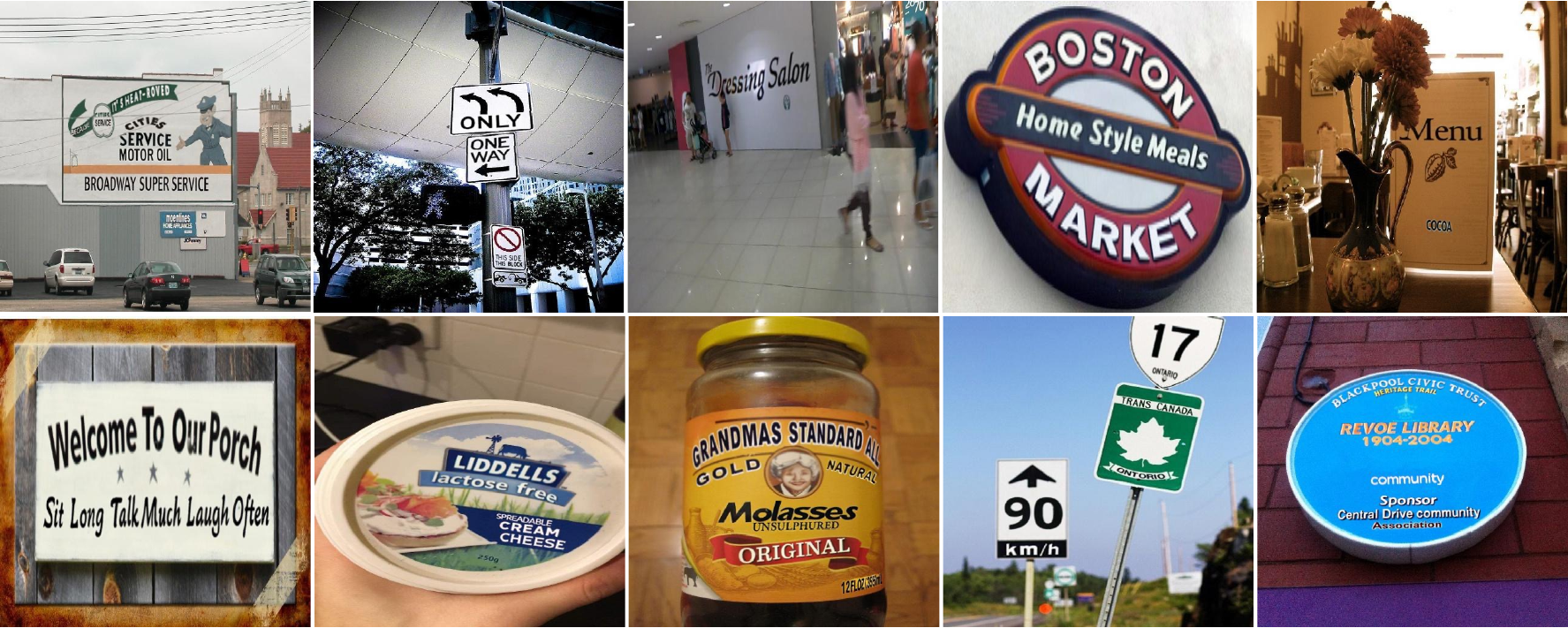}
\caption{More samples generated by LVTR on SceneVTG-benchmark.} 
\label{fig:app_fig3}
\end{figure}

\cref{fig:app_fig3} presents more LVTR generation~(\textit{local image generation} mode in Sec.~5.2 of the main text) samples on SceneVTG-benchmark. 

\subsection{TRCG generation}
\begin{figure}[ht]
\centering
\includegraphics[width=0.98\textwidth]{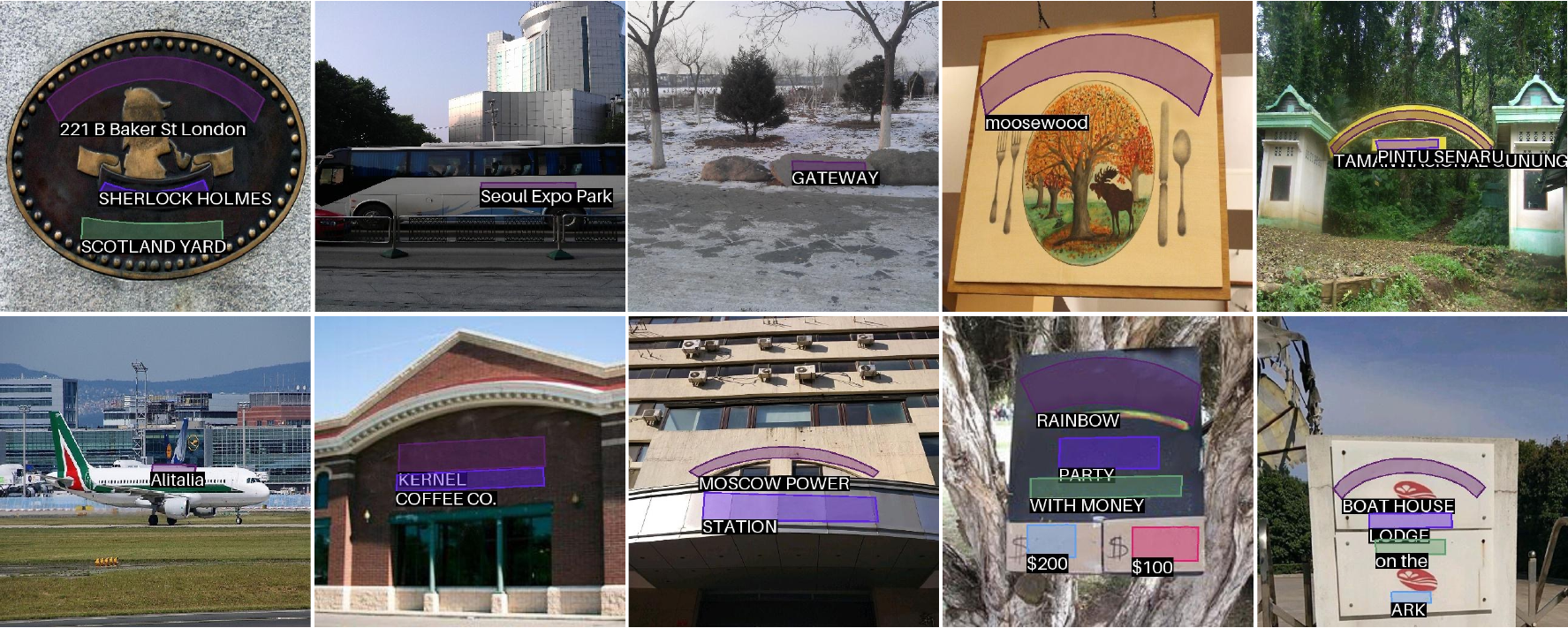}
\caption{More samples generated by TRCG on SceneVTG-benchmark.} 
\label{fig:app_fig4}
\end{figure}

\cref{fig:app_fig4} presents more TRCG generation~(same generation mode in Sec.~5.3 of the main text) samples on SceneVTG-benchmark. 

\subsection{End-to-end generation on COCO-Text dataset}
\begin{figure}[ht]
\centering
\includegraphics[width=0.98\textwidth]{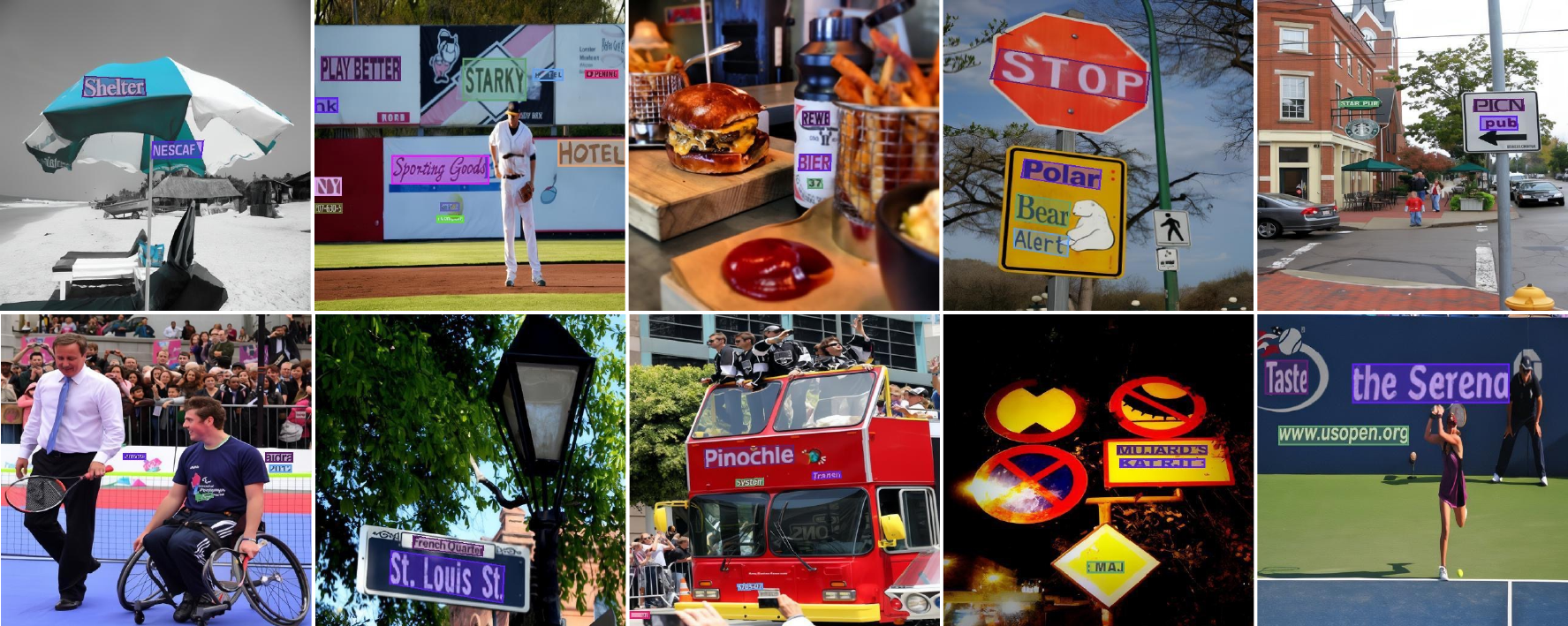}
\caption{More samples generated by end-to-end inference on COCO-Text dataset. They construct the main part of the training dataset in utility-related experiments. } 
\label{fig:app_fig5}
\end{figure}

\cref{fig:app_fig5} presents more end-to-end generation samples on COCO-Text dataset. 


\end{document}